\documentclass{article}

\usepackage{algorithm}
\usepackage{amsmath}
\usepackage{amssymb}
\usepackage{caption}
\usepackage{float}
\usepackage[margin=1.25in]{geometry}
\usepackage{graphicx}
\usepackage{paralist}
\usepackage{subcaption}

\usepackage{setspace}
\usepackage[noend]{algorithmic}

\newenvironment{Algorithm}[1][tbh]%
  {\centering
  \begin{minipage}{#1}
  \begin{algorithm}[H]}%
  {\end{algorithm}
  \end{minipage}\par
  \vspace{\belowdisplayskip}}

\newcommand{\SI}{\mbox{S-Isomap}}

\begin{document}

\title{\Large Error Metrics for Learning Reliable Manifolds from Streaming Data	}
\author{Frank Schoeneman\thanks{These authors contributed equally.}
\thanks{Department of Computer Science and Engineering, University at Buffalo, Buffalo, New York 14260}
\and
\hspace*{-.4cm}Suchismit Mahapatra\footnotemark[1] \footnotemark[2]
\and
\hspace*{-.4cm}Varun Chandola\footnotemark[2]
\and
\hspace*{-.4cm}Nils Napp\footnotemark[2]
\and
\hspace*{-.4cm}Jaroslaw Zola\footnotemark[2] \thanks{Department of Biomedical Informatics, University at Buffalo, Buffalo, New York 14260}
}
\date{}

\maketitle

\begin{abstract}
Spectral dimensionality reduction is frequently used to identify low-dimensional structure in high-dimensional data. However, learning manifolds, especially from the streaming data, is computationally and memory expensive. In this paper, we argue that a stable manifold can be learned using only a fraction of the stream, and the remaining stream can be mapped to the manifold in a significantly less costly manner. Identifying the transition point at which the manifold is stable is the key step. We present error metrics that allow us to identify the transition point for a given stream by quantitatively assessing the quality of a manifold learned using Isomap. We further propose an efficient mapping algorithm, called \SI{}, that can be used to map new samples onto the stable manifold. We describe experiments on a variety of data sets that show that the proposed approach is computationally efficient without sacrificing accuracy.
\end{abstract}

\section{Introduction}\label{sec:introduction}

Progress in science and engineering depends more than ever on our ability to analyze huge amounts of sensor and simulation data~\cite{Morik2015}. The vast majority of this data, coming from, for example, high-performance high-fidelity numerical simulations, high-resolution scientific instruments (microscopes, DNA sequencers, etc.) or Internet of Things feeds, is a result of complex non-linear processes. While these non-linear processes can be characterized by low-dimensional sub-manifolds, the actual observable data they generate is high-dimensional.

High-dimensional data is inherently difficult to explore and analyze, owing to the ``curse of dimensionality'' and ``empty space phenomenon'' that render many statistical and machine learning techniques (e.g. clustering, classification, model fitting, etc.) inadequate~\cite{Clarke2008,Scott1983}. In this context, non-linear spectral dimensionality reduction (NLSDR) has proved to be an indispensable tool~\cite{Lee2007}. However, the standard NLSDR methods, e.g. Isomap~\cite{Tenenbaum2000} or Locally Linear Embedding~\cite{Roweis2000}, have been designed for off-line or batch processing where the entire input data is available at the time of analysis. Consequently, they are computationally too expensive or impractical in cases where dimensionality reduction must be applied on a data stream.

In this paper, we address the above limitation of classic NLSDR methods and propose an alternative solution that builds on the systematic study of errors in manifold learning. We first describe a protocol to capture a ``collective error'' of the Isomap method. Then, we show how this error can be used to detect a transition point at which sufficient data has been accumulated to describe a high quality manifold, and computationally lightweight techniques can be used to efficiently map the remaining data in a stream. Our specific contributions are as follows:
\begin{enumerate}
\item We formalize a notion of {\em collective error} in Isomap and describe different strategies to quantify it using {\em Procrustes analysis}. We perform careful experimental study of the error behavior with respect to the available data using synthetic as well as real-life benchmark data. We identify properties of the error that can be used to detect when manifold becomes stable and robust to incorporating new points.
\item We propose a new efficient algorithm to incorporate streamed data into a stable Isomap manifold. The complexity of the algorithm depends only on the size of the initial stable manifold and is independent of the stream size. Thus, the algorithm is suitable for high-volume and high-throughput stream processing.
\end{enumerate}
In our approach, we focus on Isomap because of its broad adoption in scientific computing~\cite{Samudrala2015} and bio-medical research, including fMRI analysis~\cite{Thirion2004}, clustering of oncology data~\cite{Orsenigo2013}, genes and proteins expression profiling~\cite{Lee2008}, modeling of spatio-temporal relationships in data ~\cite{Jenkins2004}, and many others.

Isomap provides a simple and elegant way to estimate the intrinsic geometry of the data manifold based on a rough estimate of each data point’s neighbors on the manifold. Unfortunately, its high computational complexity means it is too expensive to use on any but relatively small data sets, and it is not suited for stream processing. Consequently, there is a significant demand for an Isomap variant that would be capable to learn a robust manifold from high-throughput data streams.

The remainder of this paper is organized as follows. In Section 2, we provide preliminary information and high level overview of NLSDR methods. In Section 3, we introduce our approach to quantify error in Isomap, and in Section 4 we report experimental results showing properties of our error measure. In Section 5, we introduce our new efficient algorithm for out-of-sample Isomap extension. We close the paper with a brief survey of related work in Section 6, and concluding remarks in Section 7.

\section{Preliminaries}\label{sec:prelim}

\begin{figure*}
  \centering
  \includegraphics[scale=0.7]{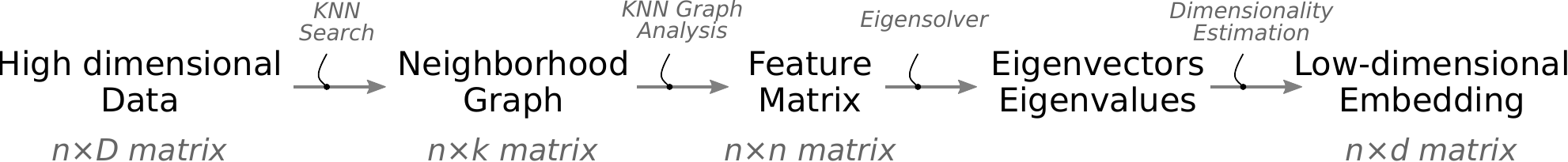}
  \caption{General non-linear spectral dimensionality reduction workflow.}\label{fig:nlsdr}
\end{figure*}

In the most general terms, the non-linear dimensionality reduction problem can be posed as follows. Given a data matrix $X = [{\bf x}_1, {\bf x}_2, \ldots, {\bf x}_n]^\top$, such that each ${\bf x}_i \in \mathbb{R}^D$, the task is to find a corresponding low-dimensional representation, ${\bf y}_i \in \mathbb{R}^d$, for each ${\bf x}_i$, where $d \ll D$. 

Non-linear spectral dimensionality reduction methods assume that the data lies along a non-linear low-dimensional manifold embedded in a high-dimensional space, and exploit the global (Isomap~\cite{Tenenbaum2000}, Minimum Volume Embedding~\cite{Weinberger2005}) or local (Locally Linear Embedding~\cite{Roweis2000}, Laplacian Eigenmaps~\cite{Belkin2001}) properties of the manifold to map each ${\bf x}_i$ to its corresponding ${\bf y}_i \in \mathbb{R}^d$.

Most NLSDR methods follow a similar series of data transformations as depicted in Figure~\ref{fig:nlsdr}. In the first step, the matrix $X$ is used to construct a neighborhood graph, where each node of the graph is connected to its $k$ nearest neighbors (KNN) in some, typically Euclidean, metric. This process involves $\mathcal{O}(n^2)$ pairwise distance computations. Next, the neighborhood graph is analyzed to construct a feature matrix. This matrix encodes properties of the input data that should be preserved during dimensionality reduction, i.e. it directly captures properties of the searched sub-manifold. For example, in the classic Isomap formulation~\cite{Tenenbaum2000}, the feature matrix stores shortest paths between each pair of points (nodes) in the neighborhood graph, which approximates the actual geodesic distance between the points. The cost of the feature matrix construction varies between different NLSDR methods, but generally falls into the range $\mathcal{O}(n)$ and $\mathcal{O}(n^3)$. To obtain the final $d$-dimensional representation of the input data, $d \ll D$, the feature matrix is factorized and the first $d$ eigenvectors become the output $Y$. The cost of this step is usually $\mathcal{O}(n^3)$.

If directly and naively applied on streaming data, NLSDR methods have to recompute the entire manifold each time a new point is extracted from a stream. This quickly becomes computationally prohibitive but guarantees the best possible quality of the learned manifold given the data. To alleviate the computational problem, landmark or general out-of-sample extension methods have been proposed (see Section 6). These techniques however either neglect manifold approximation error or remain computationally too expensive for practical applications (for example, their complexity depends directly on the stream size). Here, we propose a different strategy. Initially, we aggregate incoming data and process it using the standard Isomap. At the same time, we trace the quality of the resulting manifold with some computationally acceptable overhead. When we detect that adding new points does not improve the quality of the discovered manifold, i.e. manifold is stable, we drop the standard Isomap and proceed with a faster approximate method that is sufficient to maintain the quality of the manifold. Our approach requires two components that we describe in the following sections: a method to assess accuracy of the learned manifold, and an efficient algorithm to assimilate the remaining points from a stream.

\section{Proposed Approach to Isomap Error}\label{sec:isomap_error}

To quantitatively assess the performance of Isomap in large-scale streaming applications we first need an error metric to determine the accuracy of the learned manifold.

Error in Isomap may arise due to several reasons, including incorrect parameter selection and noisy input data. Parameter selection error may be introduced by a poor choice of neighborhood size, $k$, or selecting a sub-optimal dimensionality $d$. Error due to input data can be attributed to noise, missing or limited data entries, or a skewed, rather than uniform, sampling of the underlying manifold. Finally, there could be the intrinsic error of Isomap itself, for instance due to simplifying assumptions about manifold, limited numerical precision, etc. 

While error in Isomap has been discussed in prior work (see Section~\ref{sec:related_work}), most of these studies have focused on one particular aspect of the error. We are interested in understanding the collective error associated with Isomap as a function of the size of the input data. Here we present metrics to measure this error.

\subsection{Error Definition}

We assume that there exists a function $f: \mathbb{R}^d \rightarrow \mathbb{R}^D$ that maps each data sample ${\bf y}_i \in \mathbb{R}^d$ to ${\bf x}_i \in \mathbb{R}^D$. The goal of NLSDR is to learn the inverse mapping, $f^{-1}$, that can be used to map high-dimensional ${\bf x}_i$ to low-dimensional ${\bf y}_i$, i.e. ${\bf y}_i = f^{-1}({\bf x}_i)$.

Let the approximate mapping learned by Isomap be denoted by the inverse function $\hat{f}^{-1}$. Let $Y$ denote the data matrix containing the true low-dimensional mapping for the samples in $X$, i.e., $Y = [{\bf y}_1, {\bf y}_2,\ldots,{\bf y}_n]^\top$ and $\widehat{Y}$ denote the matrix with the approximate mapping, i.e. $\widehat{Y} = [\hat{f}^{-1}({\bf x}_1),\hat{f}^{-1}({\bf x}_2),\ldots,\hat{f}^{-1}({\bf x}_n)]^\top$. 

To measure the error between the true representation and the Isomap induced approximate representation, we leverage {\em Procrustes analysis}, which is widely used for shape analysis~\cite{Dryden1998}. The idea behind Procrustes analysis is to align two matrices, $A$ and $B$, by finding the optimal translation $t$, rotation $R$, and scaling $s$ that minimizes the Frobenius norm between the two aligned matrices, that is:
\begin{equation}
d_{Proc}(A,B) = \min_{R,t,s} \Vert sRB + t - A\Vert_F.
\label{eqn:procrustes}
\end{equation}
The above optimization problem has a closed form solution obtained by performing Singular Value Decomposition of $AB^T$~\cite{Dryden1998}.

We use the Procrustes analysis to measure the quality of the manifold approximated by Isomap in two ways:
\begin{inparaenum}[1)]
\item A direct method for comparison when low-dimensional ground truth, $Y$, is available.
\item A~reference-sample method that can be used when ground truth is absent. The asymptotic behavior of this method converges to that of the first approach.
\end{inparaenum}

\paragraph{Direct Procrustes Error}

The first approach requires the ground truth reference $Y$ for the low-dimensional manifold. When both $X$ and $Y$ are known, the error $\epsilon$ is measured using~\eqref{eqn:procrustes}, that is:
\begin{equation}
\epsilon = d_{Proc}(Y,\widehat{Y}).
\label{Error01}
\end{equation}
The error $\epsilon$ is expected to decrease with the size of the input sample $n$. Our hypothesis is that $\epsilon$ and $n$ have an inverse relationship and $\epsilon$ converges to a constant value as $n$ becomes larger. We provide an informal analysis to support this hypothesis.

For manifolds with known properties (e.g. volume, radius of curvature, etc.), the original Isomap paper~\cite{Tenenbaum2000} shows that the ratio between the approximate geodesic distance between a pair of samples computed  from the nearest neighbor graph and the true geodesic distance on the manifold can be bounded within $1 \pm \lambda$ with probability at least $1 - \mu$, for arbitrarily small values of $\lambda$ and $\mu$, if the density of the samples $\alpha$ is at least greater than $\alpha_{min}$, which is a function of $k$, $d$, $\mu$, $\lambda$, and the manifold properties ($\lambda$ is the bound on the ratio between the approximate and the true geodesic distances). Closer inspection reveals that: 
\begin{equation}
\lambda \propto \left[\frac{1}{\alpha_{min}}\right]^{\frac{1}{d}}.
\label{eqn:lambdabound}
\end{equation}

We can relate $\lambda$ and the Procrustes error $\epsilon$. In particular, using results from the Eigenvalue perturbation theory~\cite{Lancaster1985}, we note that the difference between the eigenvalues of the approximate geodesic matrix will be within $\pm \lambda$ of the eigenvalues obtained from the true geodesic matrix, which implies that $\epsilon \propto \lambda^2$. Assuming that the data is sampled uniformly and using the result from~\eqref{eqn:lambdabound}, the following holds:
\begin{equation}
\epsilon \propto \left[\frac{1}{n_{min}}\right]^{\frac{2}{d}},
\label{eqn:procrustesbound}
\end{equation}

where $n_{min}$ is the minimum number of points required to establish the error bound.

\paragraph{Reference Sample Error} 

In the absence of low-dimensional ground truth we use a reference-sample method. The reference-sample method works in the following way. Given $X$, we select $F \subset X$, a {\em reference set}, and two equal sized {\em sample sets} $R_{1}, R_{2} \subset X$. Next, we create two data sets, $D_1$ and $D_2$, such that $D_{i} = F \cup R_{i}$ for $i=1,2$. 
An approximation $\hat{f}^{-1}_{i}$ is learned for each $D_{i}$ and error is computed as the Procrustes error for the two learned approximations of the reference set $F$:
\begin{equation}
\epsilon = d_{Proc}(\hat{f}^{-1}_{1}(F), \hat{f}^{-1}_{2}(F)).
\label{Error02}
\end{equation}

As we increase the size of the sample sets $R_1$ and $R_2$ we expect the reference-sample error in~\eqref{Error02} to behave similar to the Procrustes error computed directly as in~\eqref{Error01}. In fact, we empirically show that both errors have similar asymptotic behavior on several data sets. 

\begin{figure}[t]
   \centering
   \includegraphics[]{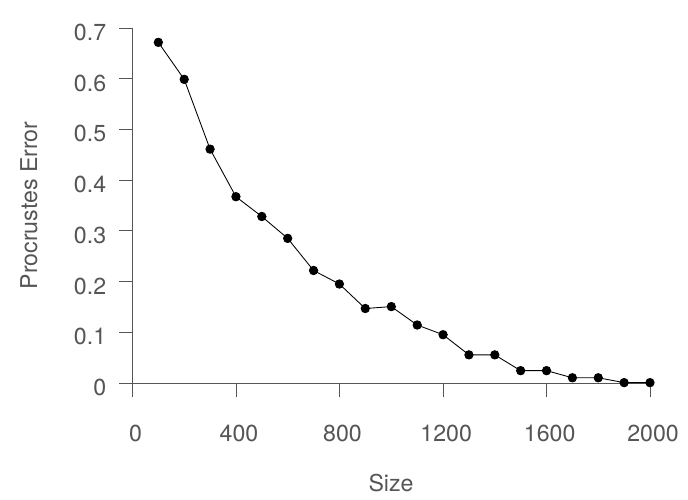}
   \caption{Isomap run on data samples of various size from the Euler Isometric Swiss Roll. The learned manifold is compared with the ground truth data using Procrustes error (\ref{Error01}).}
   \label{ISO_PE_GT}
\end{figure}

\section{Experimental Results}\label{sec:experimental_results}

We present several experiments on a variety of data sets to illustrate the behavior of error using the metrics proposed in Section \ref{sec:isomap_error}. This allows for better understanding of the asymptotic behavior of error in Isomap, required by our strategy for data stream processing. In particular, our objective is to show that the error converges after a certain number of samples are observed. 

\subsection{Test Data}

The Swiss Roll data set is typically used for evaluating manifold learning algorithms. Swiss Roll data is generated from 2-dimensional data, embedded into 3 dimensions using a non-linear function. Specifically, we have Y $\subset \mathbb{R}^2$ where: 
\begin{equation}
Y=\{(t, r) | 1\leq t \leq 3, 0\leq r \leq 1\}.
\label{SwissGT}
\end{equation}

The common approach is to generate the 3-dimensional Swiss Roll from $Y$ through the use of nonlinear functions of the form $\hat{x}(t)=\alpha t\cdot\cos{(\beta t)}$, $\hat{y}(t)=\alpha t\cdot\sin{(\beta t)}$, and $\hat{z}=r$. This is problematic due to the fact that $x(t)$, $y(t)$ are not isometric, i.e. distances between points in $Y$ are not preserved along the surface generated in $\mathbb{R}^3$. We propose a new data set to rectify this issue, the {\em Euler Isometric Swiss Roll}. The idea is to substitute the commonly used $\hat{x}(t), \hat{y}(t)$ with the equations for the {\em Euler Spiral}:
\begin{equation}
x(t) = \int_{0}^{t} \sin{(s^{2})} ds,
\label{Eulerx}
\end{equation}
\begin{equation}
y(t) = \int_{0}^{t} \cos{(s^{2})} ds.
\label{Eulery}
\end{equation}

The Euler spiral has the property that the curvature at any point is proportional to the distance from the origin. This gives constant angular acceleration along the curve thus ensuring that isometry is preserved.

In order to investigate our approach to error analysis in the absence of ground truth, we consider two benchmark data sets: the {\em MNIST handwritten digit database} and the {\em Corel Image data set}. The MNIST database is composed of 70,000 normalized, $28\times 28$ grayscale images of handwritten digits `0' to `9'. Each image is represented by a 784-dimensional vector resulting from the normalized, grayscale image. Each of the ten digits has roughly 6,000 samples. The Corel Image data set consists of 68,040 photo images from various categories. Each image is represented using 57 features.

\begin{figure}[t]
   \centering
   \includegraphics[]{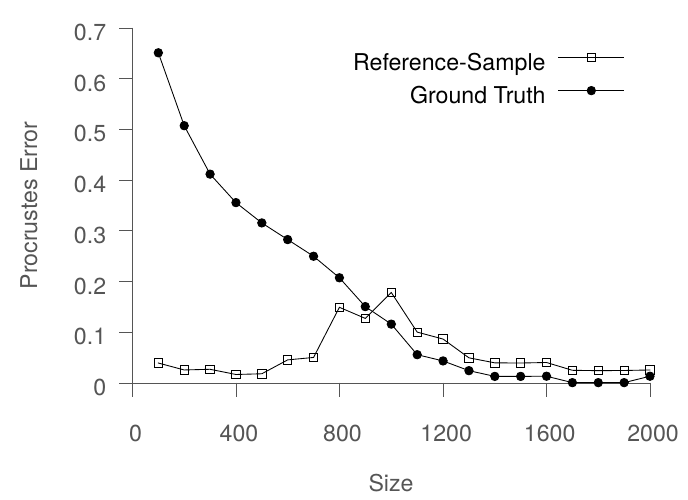}
   \caption{Isomap on two data samples of various size from the Euler Isometric Swiss Roll. First, the learned manifold is compared with the ground truth using Procrustes error~(\ref{Error01}) to illustrate the behavior of error with increasing data. Next, the shared points in both samples are compared using reference-sample error~(\ref{Error02}). In comparing (\ref{Error01}) and (\ref{Error02}) we highlight the similar asymptotic behavior of both approaches.}
   \label{ISO_SUBSAMP_PE_GT}
\end{figure}

\subsection{Results and Analysis}

First we apply Isomap to samples of data from the Swiss Roll data  set. This is to investigate the ability of Isomap to reconstruct the true low-dimensional structure under the conditions of limited data and increasing data. The obtained results are presented in Figure \ref{ISO_PE_GT}. Each experiment is performed 10 times and the mean error over all trials is reported.

Figure~\ref{ISO_PE_GT} shows that initially for smaller amounts of data the error as formulated in (\ref{Error01}) is relatively high. As more data samples are used the error first decreases rapidly and then begins to converge at approximately 1,500 samples. This suggests that the approximation may be learned within some error tolerance using 1,500 points, and that additional data points contribute no significant information. It is at this point that we wish to use more computationally lightweight, approximate methods to process the remaining samples. 

Next, we consider the reference-sample method described in Section \ref{sec:isomap_error}. In applying our sampling method to the Swiss Roll data we accomplish two goals. Like in the first experiment, we are able to investigate the quality of the Isomap approximation as data availability increases. In addition, because we have ground truth available, we are able to validate our observation that both errors asymptotically converge, in the limit of increasing data. 

In Figure \ref{ISO_SUBSAMP_PE_GT}, we present results for both approaches to error analysis. In the reference-sample method we take reference set $F$ with $|F| = 100$. We start with sample sets $R_{1}, R_{2}$ with 100 samples and increase their size by 100 samples at each step. The error between $D_{1}$ and $D_{2}$ is computed as prescribed by (\ref{Error02}). At the same time, the error for $D_{1}$ compared with ground truth coordinates (\ref{SwissGT}) is computed and plotted alongside the reference-sample error.  

\begin{figure*}[htb]
  \centering
  \begin{subfigure}{0.495\textwidth}
    \centering
    \includegraphics[]{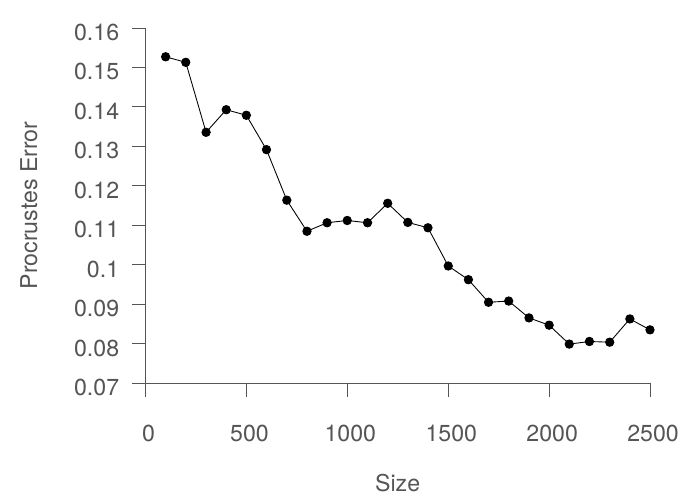}
    \caption{Digit `5' from MNIST Database}
    \label{mn5_refsamp}
  \end{subfigure}
  \begin{subfigure}{0.495\textwidth}
    \centering
    \includegraphics[]{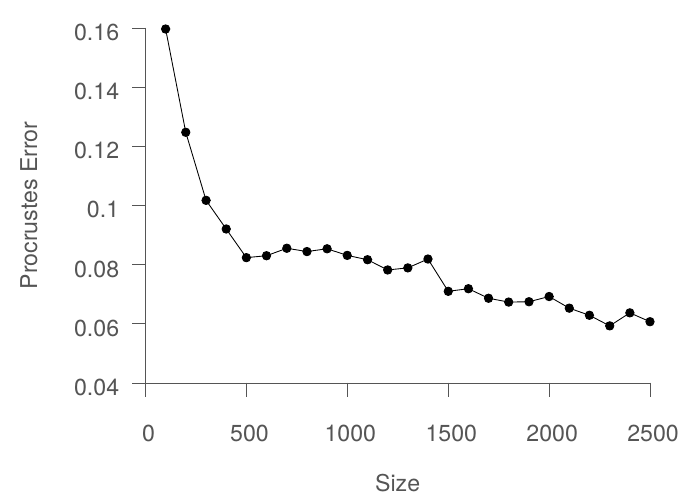}
    \caption{Digit `7' from MNIST Database}
    \label{mn7_refsamp}
  \end{subfigure}
  \begin{subfigure}{0.495\textwidth}
    \centering
    \includegraphics[]{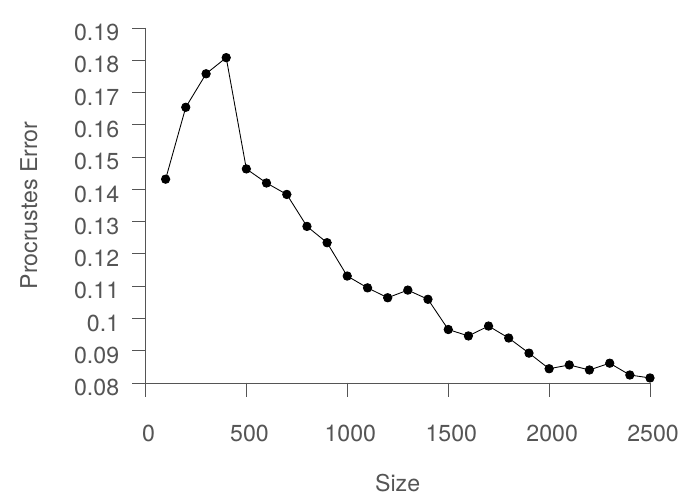}
    \caption{Digit `9' from MNIST Database}
    \label{mn9_refsamp}
  \end{subfigure}
  \begin{subfigure}{0.495\textwidth}
    \centering
    \includegraphics[]{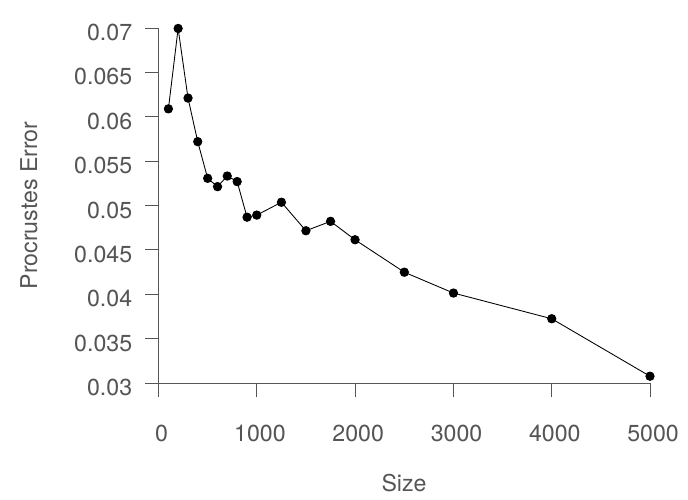}
    \caption{Corel Image Database}
    \label{corel_refsamp}
  \end{subfigure}
  \caption{Reference-sample error from~\eqref{Error02} for the benchmark data sets. The results illustrate the stabilization of the learned manifold with increasing data.}
  \label{refsamp}
\end{figure*}

Having gained an understanding of the behavior of errors on synthetic data with availability of ground truth we turn our attention to real-world data. As an example we consider samples for single digits from the MNIST database and the Corel Image data set. For each data set, we determine its intrinsic dimensionality based on residual variance and then run the reference-sample method to obtain the error plots for increasing sample size, as shown in Figure~\ref{refsamp}. 

For each data set, we observe a similar behavior to that seen for both approaches when applied to the Swiss Roll. The error decreases rapidly with additional samples. As the size of the sample sets $R_{i}$ approaches around 2,000-2,500 samples, the error begins to stabilize as both approximations become more accurate. The conclusion we draw from this is that for sample of more than $2,000$ points, a reliable manifold can be learned using both $R_1$ and $R_2$, which results in convergence of the error.

\section{Proposed Streaming Isomap Algorithm}\label{sec:stream_isomap}

To form a stable, well represented manifold, Isomap requires an adequate number of samples to learn the topology of the manifold. More samples can capture the idiosyncrasies of the manifold surface better. However, our experimental results in Section \ref{sec:experimental_results} suggest that there exists a transition point beyond which the quality of the manifold does not change significantly, even on addition of more samples. This means that once the point of transition is reached, instead of building the entire manifold in a batch fashion, we can potentially employ a computationally less expensive procedure to map the new samples arriving in the data stream. Based on this observation, we propose an out-of-sample extension method called {\em Streaming-} or {\em \SI{}} that can adequately predict the embedding of incoming samples in an efficient manner. The algorithm avoids recomputation of the entire geodesic distance matrix or its eigen decomposition, both of which are $\mathcal{O}(n^3)$. While other out-of-sample techniques exist~\cite{Bengio2004}, we demonstrate in Section~\ref{sec:streaming_analysis} that the proposed algorithm is significantly more efficient.

\begin{Algorithm}[0.45\textwidth]
    \caption{\textsc{Streaming Isomap (\SI{})}}
	\begin{algorithmic}[1]
    \setstretch{0.975}
    \REQUIRE $G_{b}$, $X_{b}$, $Y_{b}$, ${\bf x}_{s}$, $k$ 
    \ENSURE ${\bf y}_{s}$ 

    \STATE ${\bf kNN}$, ${\bf kDist}$ $\leftarrow$ KNN$($${\bf x}_{s}$, $X_{b}$, $k$$)$
    \STATE
    \FOR{$1\leq i\leq n$} 
    \STATE ${\bf g}_{i}$ $\leftarrow$ $\min_{1\leq j \leq k}\{{\bf kDist}_{j} + G_{b_{{\bf kNN}_{j},i}}\}$
    \ENDFOR \label{endfor}

    \STATE
    \STATE ${\bf c} \leftarrow \frac{1}{2}({\bar{g}}\cdot {\bf 1}_n - {\bf g} - {\bar{\bar{G}}_b}\cdot {\bf 1}_{n} + {\bar{\bf G}_{b}}$)
    \STATE ${\bf p} \leftarrow ({Y}_{b}^\top {Y}_{b})^{-1}{Y}_{b}^\top {\bf c}$
    \STATE $\hat{Y} \leftarrow $ $\lbrack {Y}_{b};{\bf p} \rbrack$
    \STATE ${\bf y}_{s} \leftarrow {\bf p} - \bar{ \hat{Y} }$ 

    \RETURN{${\bf y}_{s}$}
  \end{algorithmic}
  \label{alg:oose}
\end{Algorithm}

The algorithm assumes that a transition point has already been identified by monitoring the Isomap error using methods discussed in Section~\ref{sec:isomap_error}. Let matrix $X_b \in \mathbb{R}^{n \times D}$ denote the batch of samples encountered in the stream before the point of transition. From our previous experimental results we can assume that $n$ is relatively small compared to the remaining size of the stream. Let $X_s \in \mathbb{R}^{m \times D}$ represent the remaining part of the input data stream. Furthermore, let $G_b\in \mathbb{R}^{n \times n}$ represent the geodesic distance matrix between the samples in $X_b$, and let $Y_b \in \mathbb{R}^{n \times d}$ be the matrix containing the corresponding low-dimensional representations of the samples in $X_b$.

\subsection{Algorithm}\label{sec:streaming_algorithm}

Our proposed method is shown in Algorithm~\ref{alg:oose}. The key assumption is that because the manifold learned from $X_b$ is stable, we do not have to recompute the entire geodesic distance matrix each time a new point ${\bf x}_s$ is added. Instead, it is sufficient that we find the nearest neighbors of ${\bf x}_s$ in $X_b$ and use those to approximate geodesic distance between ${\bf x}_s$ and the remaining points in $X_b$. This step is realized in lines 1-4. Here, ${\bf kNN}$ stores indexes of the nearest neighbors of ${\bf x}_s$, ${\bf kDist}$ represents the corresponding distances, and ${\bf g}_i$ is the approximate geodesic distance between ${\bf x}_s$ and the $i$-th point in $X_b$. Given the updated geodesic distances we can obtain the low-dimensional coordinates of ${\bf x}_{s}$ using transformations similar to~\cite{Law2006}. Specifically, we match the inner product between ${\bf y}_s$ and points in $Y_{b}$ to the computed geodesics. If we denote the mean of entries of ${\bf g}$ by ${\bar g}$ and the mean of all entries of $G_b$ by ${\bar {\bar G}_{b}}$, then our desired inner product can be expressed by ${\bf c}$ as given in line 6. Here, we use ${\bf 1}_n$ to represent a vector of $n$ ones, and ${\bar{\bf G}_{b}}$ to represent the vector of row means of $G_{b}$. Then, by solving for ${\bf y}_s$, whose inner product with $Y_b$ is equal ${\bf c}$, we obtain the low-dimensional representation of ${\bf x}_s$ (lines 7-9).

\subsection{Analysis}\label{sec:streaming_analysis}

For a single point $x_s \in X_s$, the proposed \SI{} algorithm takes $\mathcal{O}(nD)$ time to compute $k$NN, $\mathcal{O}(nk)$ to compute ${\bf c}$, and $\mathcal{O}(nd^2)$ for the least-squares estimation in solving for ${\bf y}_s$, where $n$ is the size of the batch $X_b$. Thus, its runtime complexity is $\mathcal{O}(n(D + d^2+k))$ per streaming point. If we consider the cost of learning the initial manifold, the proposed \SI{} algorithm requires $\mathcal{O}(n^3 + mn(D + d^2 + k))$ time to process a stream of size $n + m$. This is significant because the cost of mapping a new sample depends only on the batch size $n$ and parameters $D, d$ and $k$. Since these are very small compared to the entire stream size, the resulting computational savings are significant. Consequently, the algorithm is well suited for high-throughput streams. In contrast, the complexity of the incremental Isomap~\cite{Law2006}, which is the other method designed for streams, scales quadratically with the size of the stream, and it can be $\mathcal{O}(iD + i^2\log(i) + i^2k)$ when inserting $i$-th sample from the stream if no initial batch is available, or $\mathcal{O}(n^2\log(n) + n^2k)$ when $X_b$ is given.

The space complexity of our algorithm is dominated by the terms $\mathcal{O}(n^2)$ and $\mathcal{O}(nd)$ due to the geodesic distance matrix, $G_b$, and the low-dimensional representation of the batch samples, $Y_b$. Thus, the space requirement does not grow with the size of the stream, which makes the algorithm appealing for handling high-volume streams.

\subsection{Experimental Results}\label{sec:streaming_results}

To validate the proposed \SI{} algorithm we perform a set of experiments to asses how accurately and efficiently it maps samples in a stream to a manifold learned from a fixed sized batch. 

\begin{figure}[t]
   \centering
   \includegraphics[]{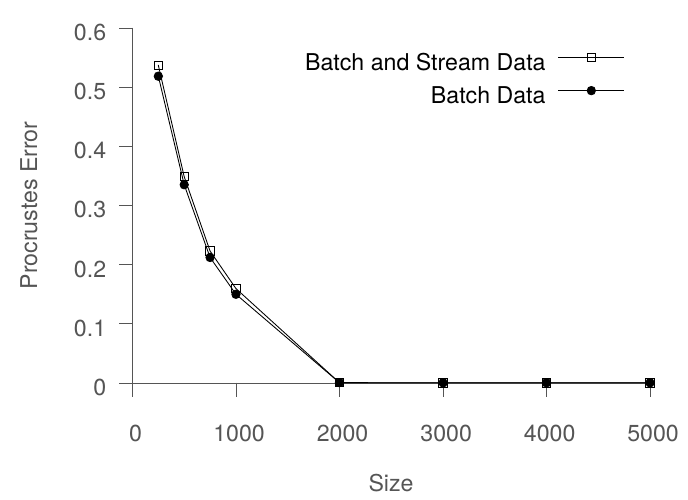}
   \caption{Effect of Procrustes error as we progressively vary the size of the batch data from Euler Isometric Swiss Roll. The Procrustes error using only the batch data set (without our proposed \SI{} algorithm) is also added as reference. The results illustrate that the error due to streaming points is low.}
   \label{str_results}
\end{figure}

\begin{figure*}[htb]
  \centering
  \begin{subfigure}{0.495\textwidth}
    \centering
    \includegraphics[]{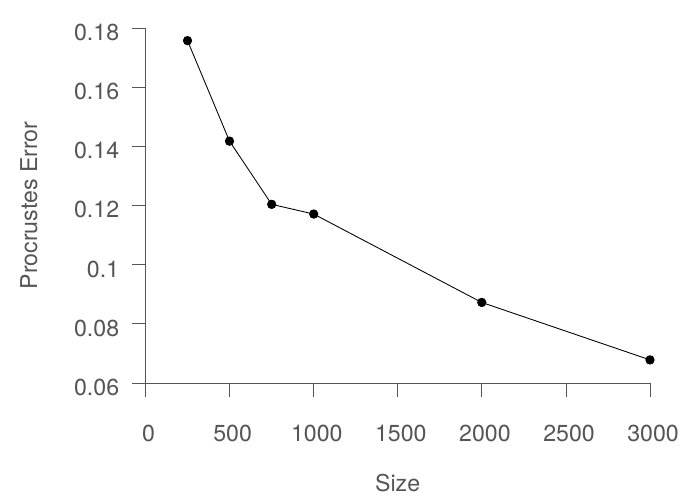}
    \caption{Digit `5' from MNIST Database}
    \label{stream_mn5}
  \end{subfigure}
  \begin{subfigure}{0.495\textwidth}
    \centering
    \includegraphics[]{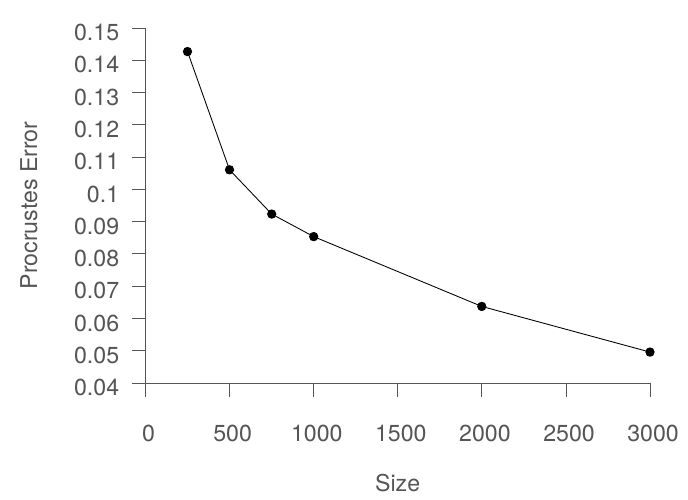}
    \caption{Digit `7' from MNIST Database}
    \label{stream_mn7}
  \end{subfigure}
  \begin{subfigure}{0.495\textwidth}
    \centering
    \includegraphics[]{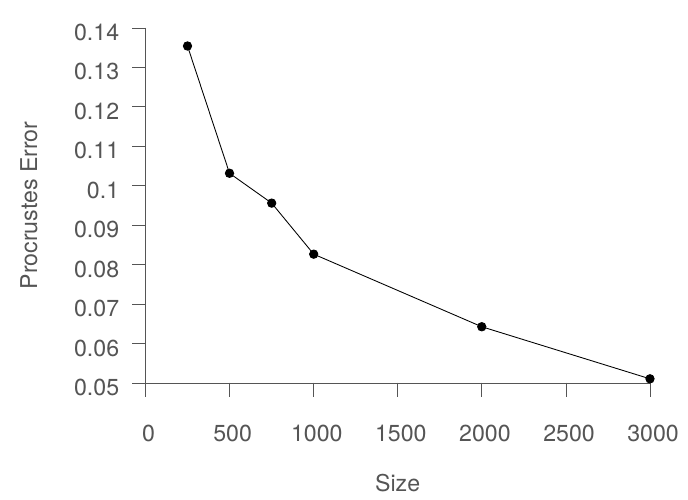}
    \caption{Digit `9' from MNIST Database}
    \label{stream_mn9}
  \end{subfigure}
  \begin{subfigure}{0.495\textwidth}
    \centering
    \includegraphics[]{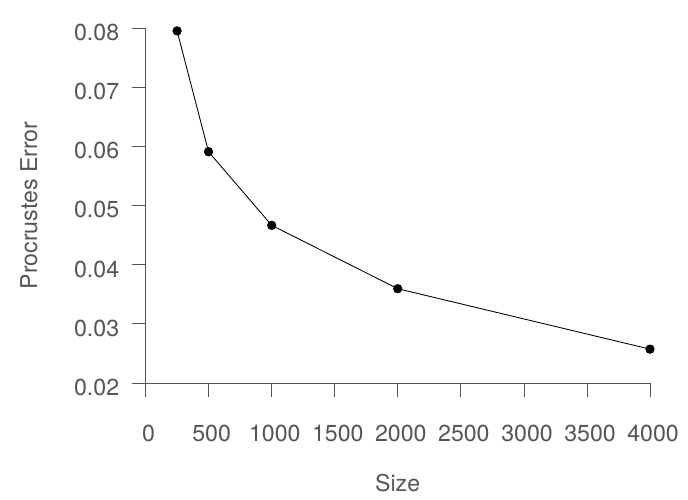}
    \caption{Corel Image Database}
    \label{stream_corel}
  \end{subfigure}
  \caption{Effect of Procrustes error as we progressively vary the size of the batch data set on benchmark data sets. The results illustrate that the error due to streaming points is very low, as well as the asymptotic behavior is almost the same.}
  \label{stream_results}
\end{figure*}
\begin{figure}
   \centering
   \includegraphics[]{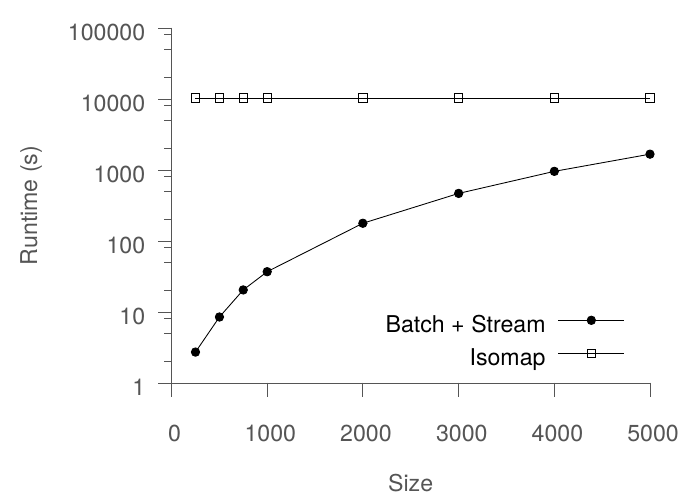}
   \caption{Timing results for our method compared to Isomap (horizontal reference line on top). The results are in log scale and demonstrate the performance gain achieved from our proposed \SI{} algorithm.\\ ~ }
   \label{str_timingresults}
\end{figure}
\begin{figure}
   \centering
   \includegraphics[]{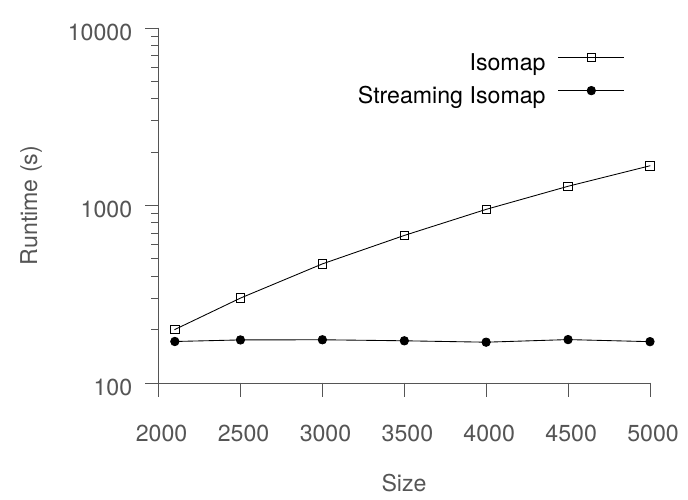}
   \caption{Timing results for \SI{} compared to Isomap for fixed batch size of 2,000 points with streaming data. The results are in log scale and demonstrate the performance gain achieved from our proposed \SI{} algorithm. }
   \label{str_timingresults1}
\end{figure}

In the first set of experiments, we measure the error in the mapping obtained using Algorithm~\ref{alg:oose} and compare it with the error in the mapping from using  standard Isomap algorithm. Using a data set $X$ of size $n + m$, we learn a manifold using a batch of size $n$ and then map the remaining $m$ samples using \SI{}. The entire mapping is compared to the ground truth using the Procrustes error. Figure~\ref{str_results} shows the results for the Isometric Swiss Roll data with $n+m=$ 10,000. Similar results are shown for the other data sets in Figure~\ref{stream_results}. For the other data sets, we use the mapping obtained by applying Isomap on full data as the ground truth. The results show that while for smaller samples the error is high, the error stabilizes to a low value after the point of transition. The error closely tracks the error for only the batch portion of the data set. This indicates that beyond the point of transition, the mapping obtained using \SI{} is as accurate as that obtained from the standard Isomap. 

To assess the computational efficiency of the \SI{} algorithm, we show the time taken by \SI{} (including the time for Isomap on the batch and \SI{} on remaining stream) in Figure~\ref{str_timingresults}. The standard Isomap timing result on all 10,000 samples is shown as the horizontal baseline for reference. We note that \SI{} is able to map the samples in the stream much more efficiently than Isomap on the entire data. Overall, Figures~\ref{str_results} and~\ref{str_timingresults} show that the proposed method is able to map samples in a stream in a highly efficient manner without sacrificing the quality of the manifold.

To understand how \SI{} would behave in a truly streaming mode we conduct a second timing experiment. Here, we fix the size of the batch to the estimated switching point (2,000 for the Isometric Swiss Roll) and then progressively add samples to the stream and process it with \SI{}. We measure the cumulative time taken by \SI{} to process the remainder of the stream, for different sizes. Figure~\ref{str_timingresults1} shows the times compared with the runtime for running Isomap on the aggregated batch and stream data. As previously, the cumulative time of running \SI{} scales linearly with the size of the stream. This further confirms efficiency of the method.

\section{Related Work}\label{sec:related_work}

Given the high computational complexity of Isomap, variants of Isomap, such as Landmark Isomap~\cite{Silva2002} and out-of-sample extension techniques~\cite{Bengio2004}, have been proposed as a computationally viable alternative. Both of these methods either use a smaller set of landmark points or approximations to avoid performing the costly eigen decomposition on the $n\times n$ geodesic distance matrix, where $n$ is the number of points in the entire data set. However, they still require computing the full geodesic distance matrix which is $\mathcal{O}(r n^{3})$, where $r$ is the diameter of the embedded $k$NN graph. The Incremental Isomap algorithm~\cite{Law2006} avoids both eigen decomposition and a recreation of the geodesic distance matrix. However, it requires updates to the geodesic distance matrix that incurs a significant cost, as discussed in Section~\ref{sec:streaming_analysis}. Consequently, the method is unsuitable \mbox{for~the~streaming~setting}.

Errors in Isomap have been discussed in prior work~\cite{Lee2007}, but those studies have been typically in regards to selection of parameters. For example, Samko et al.~\cite{Samko2006} proposed measuring a simple manifold embedding error for a range of $k$ to find the best choice~of~$k$. Similarly, an approach based on the $k$-edge disjoint minimal spanning tree algorithm has been proposed to construct a neighborhood graph with connectivity guarantees~\cite{Yang2005}. In the same spirit, several strategies to assess intrinsic manifold dimension $d$ are available~\cite{Lee2007}. However, to the best of our knowledge there is no prior work in defining and understanding the behavior of error that persists even with the selection of optimal parameters. We address this error in taking an abstract view of Isomap, providing a protocol for measuring collective error, and understanding its behavior. In doing so we identify the optimal point where we may switch from exact to lightweight methods.

\section{Conclusions}\label{sec:conclusion}

The error in Isomap approaches $0$ as the density of sampled data tends to infinity. However, in practical settings, we can expect that after a certain level of sampling the error does not change significantly. In other words, the learned manifold becomes stable if the sample size reaches a certain threshold, under the assumption of uniform sampling. In this paper, we have presented the error metrics that can be used to empirically observe when the manifold becomes stable. In particular, the reference-sample metric is appealing because it can assess the manifold quality even when ground truth data is unavailable.

Equipped with the knowledge of the point of transition, we have presented a streaming algorithm, \SI{}, that can be used to efficiently map new data samples to the stable manifold, instead of performing costly updates. The fact that the cost of mapping new samples in \SI{} depends only on the size of the initial batch used to generate the stable Isomap, and is independent of the size of the stream itself, makes it a very powerful tool to process massive streams of data.

\section{Acknowledgment}\label{sec:acknowledgement}

Authors  wish  to  acknowledge  support  provided  by  the Center  for  Computational  Research  at  the  University  at Buffalo.
This work has been supported in part by the NSF grants CNS-1409551 and IIS-1651475.

\bibliography{references}{}
\bibliographystyle{plain}

\end{document}